\documentclass{article}
\usepackage{ijcai20}

\usepackage{times}
\usepackage{soul}
\usepackage{url}
\usepackage[hidelinks]{hyperref}
\usepackage[utf8]{inputenc}
\usepackage[small]{caption}
\usepackage{graphicx}
\usepackage{amsmath}
\usepackage{amsthm}
\usepackage{booktabs}
\usepackage{algorithm}
\usepackage{algorithmic}
\usepackage{comment}
\title{Fact Check: Analyzing Financial Events from Multilingual News Sources}

\author{
Linyi Yang$^1$\and
Tin Lok James Ng$^2$\and
Barry Smyth$^1$\And
Ruihai Dong$^1$
\affiliations
$^1$Insight Centre, University College Dublin\\
$^2$Trinity College Dublin
\emails
\{linyi.yang, barry.smyth, ruihai.dong\}@insight-centre.org,
jamesng@tcd.ie.
}

\begin{document}

\maketitle

\begin{abstract}
The explosion in the sheer magnitude and complexity of financial news data in recent years makes it increasingly challenging for investment analysts to extract valuable insights and perform analysis. We propose FactCheck in finance, a web-based news aggregator with deep learning models, to provide analysts with a holistic view of important financial events from multilingual news sources and extract events using an unsupervised clustering method. A web interface is provided to examine the credibility of news articles using a transformer-based fact checker. The performance of the fact checker is evaluated using a dataset related to merger and acquisition (M\&A) events and is shown to outperform several strong baselines. 
\end{abstract}

\section{Introduction}
A tremendous amount of financial news sources are available online nowadays. The large weave of information generated by these news sources can be overwhelming and confusing for financial investors, which makes the discrimination of useful information out of tons of worthless or repetitive data a challenging and tedious task. While an investor can rely on a particular news source to learn about current financial events, the coverage of the events will not be comprehensive and may be constrained to the geographic location of the news source. On the other hand, the use of a financial news aggregator can provide investors with a coverage of the same financial events from multiple news sources. 

Recently, the use of deep learning and natural language processing (NLP) in the analysis of unstructured data has attracted much research interests in the financial domain \cite{Ding14,yang2018explainable,yang2020maec,Yang20}. Motivated by the impressive success in the applications of these methods, we propose an event-based financial news aggregator along with a novel, transformer-based architecture for assessing the validity of events. While various general purpose and domain-specific news aggregators are available, our objective is to develop a financial news aggregator which gathers relevant news information from multiple sources in different languages. Unlike many general purpose aggregators\cite{zhang2019tanbih}, the news aggregator we introduce identifies and constructs actual events that are being discussed in the articles from various news sources, which aims to provide investors with a holistic view of important events. Figure 1 describes an example of cross-lingual linking of news articles related to accounting scandal of Luckin Coffee which caused its share price to tumble as much as 83\%.

The accuracy of each constructed event is assessed using a transformer-based fact checker. Existing fact checking of news events in the financial domain is typically performed by human experts\footnote{\url{http://www.factcheck.org/}}\footnote{\url{http://www.snopes.com/}}, which is constrained by the experience of the journalists and lacks scalability. The transformer-based fact checker overcomes these shortcomings and achieved reliable results,  as demonstrated in our case study.

\begin{figure}[t]
\centering
{%
\includegraphics[width = \linewidth]{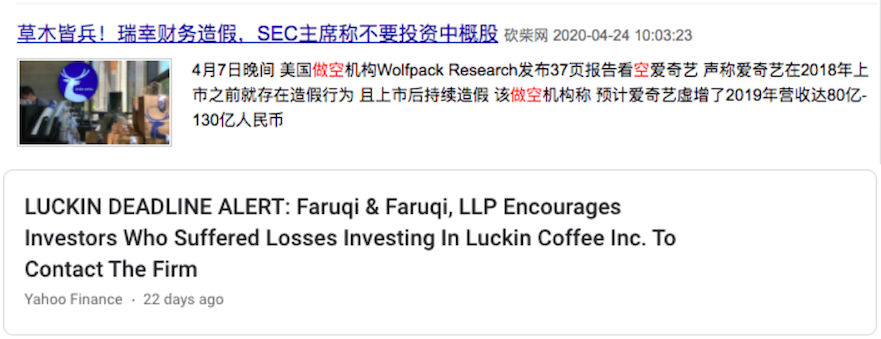}
}
\caption{News articles related to the accounting scandal of Lucking Coffee in both English and Chinese language.}

\end{figure}

\section{Detailed Implementations}
In this section, we describe our data first. Then, we introduce the process of building our system based on AllenNLP\footnote{\url{https://github.com/allenai/allennlp}} with a micro web framework, Flask. Lastly, a novel transformer-based prediction model will be discussed.

The architecture of our system is sketched in Figure 2. Our system mainly contains two components: News Aggregator, and Fact Checker. In addition, we have translated all news in bilingual of English and Simplified Chinese by a online API-based translator\footnote{\url{https://fanyi-api.baidu.com/}}. The primary technical contributions include a event-based financial news aggregator and a fact checker based on a novel, transformer-based architecture. 

\begin{figure}[t]
\centering
{%
\includegraphics[width = \linewidth]{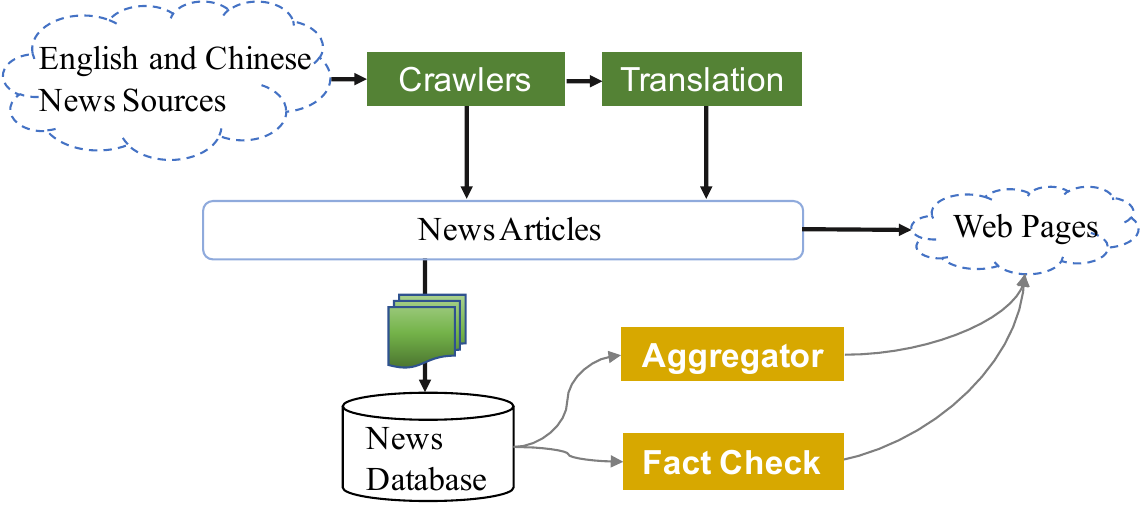}
}
\caption{The overall architecture of our system.}

\end{figure}

\subsection{Data}
We collected news articles published between January 1st, 2020 and April 22nd, 2020 from Reuters, Fox and Sina. The resulting dataset contains a total of 183,991 news articles with 116,390 in English and 67,601 in Chinese. 

\subsection{News Aggregator}
Our website provides both English and Chinese language versions. Non-English news articles are first translated into English, and event extraction is subsequently performed using an unsupervised news clustering model proposed by \cite{ribeiro2017unsupervised}. More specifically, hashtags, which correspond to the events that are mentioned in the news article, are assigned to each article. The result of each search query depends on both the hashtags and the content of the news articles, and news articles mentioning the same events in different languages will be jointly displayed (as shown in Figure 3).

\subsection{Fact Checker}
In order to assess the credibility of each news article, we adopt a transformer-based fact checker. We leverage the adversarial training strategy, namely, projected gradient descent (PGD) \cite{madry2018towards} for training the transformer. The adversarial training is implemented on WWM-BERT, whereby perturbations are added to the input embedding to improve the robustness of the model.

As a case study, we train a transformer-based classifier to predict the outcome of an merger and acquisition (M\&A) event \cite{yang2020generating} using relevant news articles. A dataset which contains news articles related to 14,539 M\&A deals between January 1st, 2007 and August 12th, 2019 were collected from Zephyr. After preprocessing, the final dataset contains 4,098 instances which were further divided into a training, validation and test set.

\begin{figure}[t]
\centering
{%
\includegraphics[width = \linewidth]{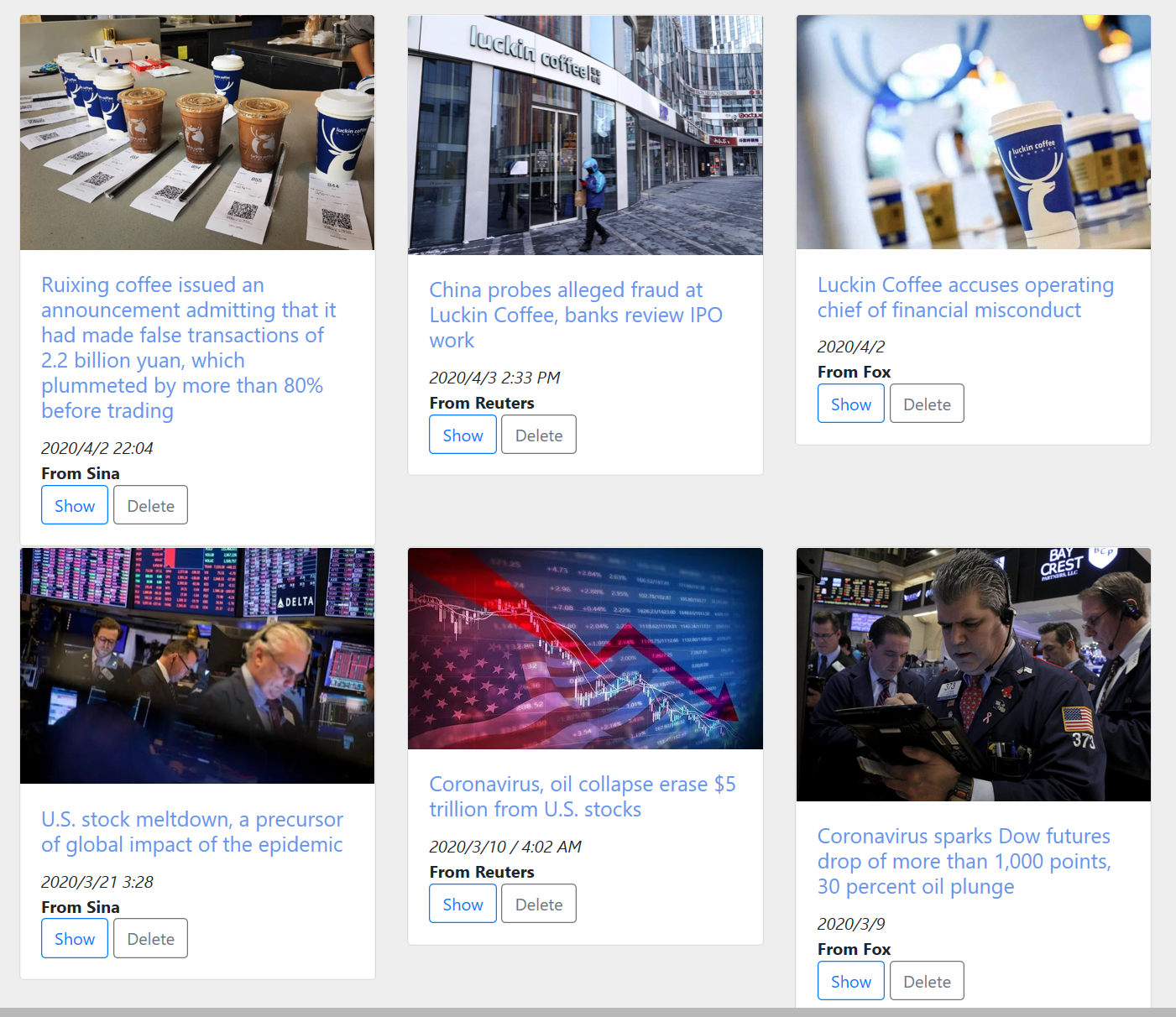}
}
\caption{The page of the news aggregator. Each column represents the news from each source, while each row demonstrates the different reports of a given event.}
\end{figure}

\section{A Case Study on M\&A Predictions}
We evaluate the performance of the binary classification task described above. The adversarial transformer is compared with several deep learning based methods and an original transformer where the results are summarized in Table 1. The proposed model achieves significant improvement over strong baselines in terms of MCC, accuracy and F1 scores. Additionally, the improvement over the original transformer is statistically significant according to an independent t-test (with a p-value of 0.014). The fine-tuned transformer model can be leveraged to assess the credibility of news articles for a wide range of financial events.

\begin{table}[t]
\centering
\begin{tabular}{llll}
\hline
Evaluation & MCC & Acc & F1 \\ 
\hline
Random Guess & 0.013 & 0.510 & 0.462 \\
\textbf{Baselines} &  &  &  \\
CNN-Text & 0.729 & 0.848 & 0.847 \\
BiGRU & 0.734 & 0.836 & 0.849 \\
HAN & 0.742 & 0.848 & 0.853 \\
BERT-WWM & 0.751 & 0.874 & 0.879 \\
\textbf{Ours} &  &  &  \\
\textit{BERT-WWM+adv.} &\textit{\textbf{0.788}} & \textit{\textbf{0.894}} & \textit{\textbf{0.894}} \\ \hline
\end{tabular}
\label{tab:plain}
\caption{``MCC" is the score of Matthews Correlation Coefficient and ``adv" is short for the adversarial training by using Projected Gradient Descent (PGD).}
\end{table}
\section{Discussion}
A website for aggregating financial news articles and assessing the credibility of news articles is described in this paper. The developed adversarial transformer-based fact checker is shown to outperform a range of strong baselines in terms of predicting the outcomes of M\&A deals. The trained transformer is readily applicable to predict the outcomes of various financial events. 

For future work, the developed system can be expanded to include additional news sources in multiple languages. Furthermore, the pre-trained transformer can be transferred to a wide range of financial event prediction problems.

With the development of Large-scale Language Models, the fact-check ability of models is re-defined and can be measured in different settings, such as the previous attempts to identify pitfalls in current natural language understanding systems using counterfactual explanations \cite{yang2020generating,yang2021exploring}, and design unified OOD benchmarks \cite{yang2023glue,wang2023robustness} as well as causality-inspired evaluation tools \cite{yang2023learning}. Also, human-in-the-loop and rationale-based methods have received decent focus in recent \cite{yang2022numhtml,wang2022usb,yang2022rationale,yang2022factmix}. Besides, the evaluation of the fact-check ability of LLMs \cite{chang2023survey,zhu2023promptbench,wang2023pandalm,yidong2023pandalm}, as well as hallucinations, need to be paid more attention to.

\bibliographystyle{named}
\bibliography{ijcai20}

\begin{thebibliography}{}

\bibitem[\protect\citeauthoryear{Chang \bgroup \em et al.\egroup
  }{2023}]{chang2023survey}
Yupeng Chang, Xu~Wang, Jindong Wang, Yuan Wu, Kaijie Zhu, Hao Chen, Linyi Yang,
  Xiaoyuan Yi, Cunxiang Wang, Yidong Wang, et~al.
\newblock A survey on evaluation of large language models.
\newblock {\em arXiv preprint arXiv:2307.03109}, 2023.

\bibitem[\protect\citeauthoryear{Ding \bgroup \em et al.\egroup
  }{2014}]{Ding14}
Xiao Ding, Yue Zhang, Ting Liu, and Junwen Duan.
\newblock Using structured events to predict stock price movement: An empirical
  investigation.
\newblock In {\em Proceedings of the 2014 Conference on Empirical Methods in
  Natural Language Processing (EMNLP)}, pages 1415--1425, 2014.

\bibitem[\protect\citeauthoryear{Madry \bgroup \em et al.\egroup
  }{2018}]{madry2018towards}
Aleksander Madry, Aleksandar Makelov, Ludwig Schmidt, Dimitris Tsipras, and
  Adrian Vladu.
\newblock Towards deep learning models resistant to adversarial attacks.
\newblock In {\em Proceedings of the International Conference on Learning
  Representations (ICLR)}, 2018.

\bibitem[\protect\citeauthoryear{Niu \bgroup \em et al.\egroup
  }{2023}]{yang2023learning}
Yingjie Niu, Linyi Yang, Ruihai Dong, and Yue Zhang.
\newblock Learning to generalize for cross-domain qa.
\newblock In {\em Proceedings of the 60th Annual Meeting of the Association for
  Computational Linguistics (Volume 1: Long Papers)}, 2023.

\bibitem[\protect\citeauthoryear{Ribeiro \bgroup \em et al.\egroup
  }{2017}]{ribeiro2017unsupervised}
Swen Ribeiro, Olivier Ferret, and Xavier Tannier.
\newblock Unsupervised event clustering and aggregation from newswire and web
  articles.
\newblock In {\em Proceedings of the 2017 EMNLP Workshop: Natural Language
  Processing meets Journalism}, pages 62--67, 2017.

\bibitem[\protect\citeauthoryear{Wang \bgroup \em et al.\egroup
  }{2022}]{wang2022usb}
Yidong Wang, Hao Chen, Yue Fan, Wang Sun, Ran Tao, Wenxin Hou, Renjie Wang,
  Linyi Yang, Zhi Zhou, Lan-Zhe Guo, et~al.
\newblock Usb: A unified semi-supervised learning benchmark for classification.
\newblock {\em Advances in Neural Information Processing Systems},
  35:3938--3961, 2022.

\bibitem[\protect\citeauthoryear{Wang \bgroup \em et al.\egroup
  }{2023a}]{wang2023robustness}
Jindong Wang, Xixu Hu, Wenxin Hou, Hao Chen, Runkai Zheng, Yidong Wang, Linyi
  Yang, Haojun Huang, Wei Ye, Xiubo Geng, et~al.
\newblock On the robustness of chatgpt: An adversarial and out-of-distribution
  perspective.
\newblock {\em arXiv preprint arXiv:2302.12095}, 2023.

\bibitem[\protect\citeauthoryear{Wang \bgroup \em et al.\egroup
  }{2023b}]{wang2023pandalm}
Yidong Wang, Zhuohao Yu, Zhengran Zeng, Linyi Yang, Cunxiang Wang, Hao Chen,
  Chaoya Jiang, Rui Xie, Jindong Wang, Xing Xie, et~al.
\newblock Pandalm: An automatic evaluation benchmark for llm instruction tuning
  optimization.
\newblock {\em arXiv preprint arXiv:2306.05087}, 2023.

\bibitem[\protect\citeauthoryear{Yang \bgroup \em et al.\egroup
  }{2018}]{yang2018explainable}
Linyi Yang, Zheng Zhang, Su~Xiong, Lirui Wei, James Ng, Lina Xu, and Ruihai
  Dong.
\newblock Explainable text-driven neural network for stock prediction.
\newblock In {\em 2018 5th IEEE International Conference on Cloud Computing and
  Intelligence Systems (CCIS)}, pages 441--445. IEEE, 2018.

\bibitem[\protect\citeauthoryear{Yang \bgroup \em et al.\egroup
  }{2020a}]{yang2020generating}
Linyi Yang, Eoin Kenny, Tin Lok~James Ng, Yi~Yang, Barry Smyth, and Ruihai
  Dong.
\newblock Generating plausible counterfactual explanations for deep
  transformers in financial text classification.
\newblock In {\em Proceedings of the 28th International Conference on
  Computational Linguistics}, pages 6150--6160, 2020.

\bibitem[\protect\citeauthoryear{Yang \bgroup \em et al.\egroup
  }{2020b}]{yang2020maec}
Linyi Yang, Jiazheng Li, Barry Smyth, and Ruihai Dong.
\newblock Maec: A multimodal aligned earnings conference call dataset for
  financial risk prediction.
\newblock In {\em Proceedings of the 29th ACM International Conference on
  Information \& Knowledge Management}, pages 3063--3070, 2020.

\bibitem[\protect\citeauthoryear{Yang \bgroup \em et al.\egroup
  }{2020c}]{Yang20}
Linyi Yang, Tin Lok~James Ng, Barry Smyth, and Ruihai Dong.
\newblock Html: Hierarchical transformer-based multi-task learning for
  volatility prediction.
\newblock In {\em Proceedings of The Web Conference 2020}, WWW’20, page
  441–451, 2020.

\bibitem[\protect\citeauthoryear{Yang \bgroup \em et al.\egroup
  }{2021}]{yang2021exploring}
Linyi Yang, Jiazheng Li, P{\'a}draig Cunningham, Yue Zhang, Barry Smyth, and
  Ruihai Dong.
\newblock Exploring the efficacy of automatically generated counterfactuals for
  sentiment analysis.
\newblock In {\em Proceedings of the 59th Annual Meeting of the Association for
  Computational Linguistics and the 11th International Joint Conference on
  Natural Language Processing (Volume 1: Long Papers)}, pages 306--316, 2021.

\bibitem[\protect\citeauthoryear{Yang \bgroup \em et al.\egroup
  }{2022a}]{yang2022numhtml}
Linyi Yang, Jiazheng Li, Ruihai Dong, Yue Zhang, and Barry Smyth.
\newblock Numhtml: Numeric-oriented hierarchical transformer model for
  multi-task financial forecasting.
\newblock In {\em Proceedings of the AAAI Conference on Artificial
  Intelligence}, volume~36, pages 11604--11612, 2022.

\bibitem[\protect\citeauthoryear{Yang \bgroup \em et al.\egroup
  }{2022b}]{yang2022rationale}
Linyi Yang, Jinghui Lu, Brian Mac~Namee, and Yue Zhang.
\newblock A rationale-centric framework for human-in-the-loop machine learning.
\newblock In {\em Proceedings of the 60th Annual Meeting of the Association for
  Computational Linguistics (Volume 1: Long Papers)}, 2022.

\bibitem[\protect\citeauthoryear{Yang \bgroup \em et al.\egroup
  }{2022c}]{yang2022factmix}
Linyi Yang, Lifan Yuan, Leyang Cui, Wenyang Gao, and Yue Zhang.
\newblock Factmix: Using a few labeled in-domain examples to generalize to
  cross-domain named entity recognition.
\newblock In {\em Proceedings of the 29th International Conference on
  Computational Linguistics}, pages 5360--5371, 2022.

\bibitem[\protect\citeauthoryear{Yang \bgroup \em et al.\egroup
  }{2023}]{yang2023glue}
Linyi Yang, Shuibai Zhang, Libo Qin, Yafu Li, Yidong Wang, Hanmeng Liu, Jindong
  Wang, Xing Xie, and Yue Zhang.
\newblock Glue-x: Evaluating natural language understanding models from an
  out-of-distribution generalization perspective.
\newblock In {\em Proceedings of the 61st Annual Meeting of the Association for
  Computational Linguistics (Volume 1: Long Papers)}, 2023.

\bibitem[\protect\citeauthoryear{Yidong \bgroup \em et al.\egroup
  }{2023}]{yidong2023pandalm}
Wang Yidong, Yu~Zhuohao, Zeng Zhengran, Yang Linyi, Heng Qiang, Wang Cunxiang,
  Chen Hao, Jiang Chaoya, Xie Rui, Wang Jindong, et~al.
\newblock Pandalm: Reproducible and automated language model assessment, 2023.

\bibitem[\protect\citeauthoryear{Zhang \bgroup \em et al.\egroup
  }{2019}]{zhang2019tanbih}
Yifan Zhang, Giovanni Da~San~Martino, Alberto Barr{\'o}n-Cedeno, Salvatore
  Romeo, Jisun An, Haewoon Kwak, Todor Staykovski, Israa Jaradat, Georgi
  Karadzhov, Ramy Baly, et~al.
\newblock Tanbih: Get to know what you are reading.
\newblock {\em EMNLP-IJCNLP 2019}, pages 223--228, 2019.

\bibitem[\protect\citeauthoryear{Zhu \bgroup \em et al.\egroup
  }{2023}]{zhu2023promptbench}
Kaijie Zhu, Jindong Wang, Jiaheng Zhou, Zichen Wang, Hao Chen, Yidong Wang,
  Linyi Yang, Wei Ye, Neil~Zhenqiang Gong, Yue Zhang, et~al.
\newblock Promptbench: Towards evaluating the robustness of large language
  models on adversarial prompts.
\newblock {\em arXiv preprint arXiv:2306.04528}, 2023.

\end{thebibliography}

\end{document}